# Large Language Models Understanding: an Inherent Ambiguity Barrier

Daniel N. Nissani (Nissensohn)

dnissani@post.bgu.ac.il

ABSTRACT

A lively ongoing debate is taking place, since the extraordinary emergence of Large Language Models (LLMs) with regards to their capability to understand the world and capture the meaning of the dialogues in which they are involved. Arguments and counter-arguments have been proposed based upon thought experiments, anecdotal conversations between LLMs and humans, statistical linguistic analysis, philosophical considerations, and more. In this brief paper we present a counter-argument based upon a thought experiment and semi-formal considerations leading to an inherent ambiguity barrier which prevents LLMs from having any understanding of what their amazingly fluent dialogues mean.

1. INTRODUCTION

The emergence of Large Language Models (LLMs in the sequel) and the unanimous recognition of their extraordinary and quite surprising fluency immediately generated a still ongoing debate within the research community (as well as within the layman public) regarding these models intelligence in general and their language and world understanding in particular.

On the pro side of this debate, arguing that LLMs have at least some understanding of the meaning of the words they are exchanging we may find, amongst others, works regarding physical properties of the world (e.g. Abdou et al., 2021); analysis of anecdotal conversations between humans and LLMs (e.g. Aguera y Arcas, 2021); thought and real life experiments (e.g. Sogaard, 2023); interviews and essays (Hinton, 2024; Manning, 2022).

On the con side we find again thought experiments (e.g. Bender and Koller, 2020); essays (Browning and LeCun, 2022; Marcus, 2022; Bisk et al, 2020; Mahowald et al., 2024); formal arguments (e.g. Merrill et al, 2021); statistical linguistics analysis results (e.g. Niven and Kao, 2019), and more.

Surveys describing the 'state of the debate' have also been published (Mitchell and Krakauer, 2023; Michael et al., 2022) which indicate an approximate 50/50 opinion split amongst the research community members.

Starting from about three decades ago, in a sequence of pioneering experiments (Fried et al., 1997, Kreiman et al., 2000; Quiroga et al., 2005) uncovered a region in human brains which function as a 'center of abstract concepts': neural cells that selectively and strongly respond to the presence of stimuli of various modalities which invoke a specific abstract concept (such as the now famous Jennifer Aniston cells). More recent results (Bausch et al., 2021) provide evidence for additional neural cells which encode relations (such as "Bigger?", "More expensive?", etc.) between pairs of concepts.

These findings have inspired us to propose, in this brief paper, a simple but useful language model, presented in Section 2 which we may apply to humans involved in communication. We then show in Section 3 why this same language model cannot be applied to communicating LLMs (assuming their current state-of-the-art architectures and training protocols). This finally leads through a thought experiment and a semi-formal argument to an inherent ambiguity barrier which prevents LLMs from referring words to definite abstract concepts, and thus from understanding the meaning of their dialogues. We discuss and summarize our results in Section 4.

2. A LANGUAGE MODEL

We adopt herein a model where a *language* is a pair of mappings, L and its inverse $L^{-1}$, between a finite set of *concepts abstract representations* K = {$k_i$} (*concepts* for short in the sequel) and a *vocabulary*, i.e. a finite set of *words* of same size W = {$w_i$} with i = 1, 2, … N = |W| = |K|:

L :     K     →     W                                                                                    (1a), and

$L^{-1}$:     W     →     K                                                                                    (1b)

The concepts set K and its associated entities contain all that is needed for an agent to 'feel like it understands' the world. These associated entities include the vast and intricate set of relations amongst the K elements themselves and amongst them and the 'grounded' physical world. The maps L and $L^{-1}$ may look trivial but they are not. In particular, the words of set W, in absence of L and $L^{-1}$, are devoid of any sense of meaning: no human, lacking the maps L and $L^{-1}$, will understand what W elements mean, as anyone listening to a foreign language may testify. K is then a simple model of the fore-mentioned abstract 'concepts center' (Fried et al., 1997). It is not required for our purpose to make these statements less vague than this.

As an indirect result of the meanings of the elements of K and of the fore mentioned set of relations (and influenced also by the language syntax rules), the next word in a *sentence* (or the first word in a new sentence), denoted $w_i$, turns out to be stochastically constrained by the conditional probabilities matrix

$$\Pr\{ W = w_i \mid c_j \in C \} \equiv P_{ij} \qquad (2)$$

where $c_j$ is a *context* which conditions the random variable W realization, i.e. the word $w_i$ (note our slight abuse of notation: W may represent here either a r.v. or our vocabulary set). This context $c_j = [s_{j1}, s_{j2}, ….]$, element of the huge set of contexts C, is made up of a finite sequence of syntactically and semantically valid sentences, and each sentence $s_{jk} = [w_{jk1}, w_{jk2}, … ]$, is made up of a finite sequence of words, expressed as vocal utterances or textual symbols.

We ignore in the sequel, for simplicity, synonyms, homonyms and other vocabulary nuisances. This simplification makes L (and $L^{-1}$) a bijective function as expressed by Eq. 1b above.

The elements $w_i$ of the vocabulary and the language syntax rules are defined by convention amongst agents of the same community, and are quite arbitrary. Intelligent and communicative agents learn L and $L^{-1}$ along with the sets K and W (and the conditionals matrix $P_{ij}$). Intelligent but non-communicative agents learn only K.

A schematic embodiment of our model is depicted in Figure 1.

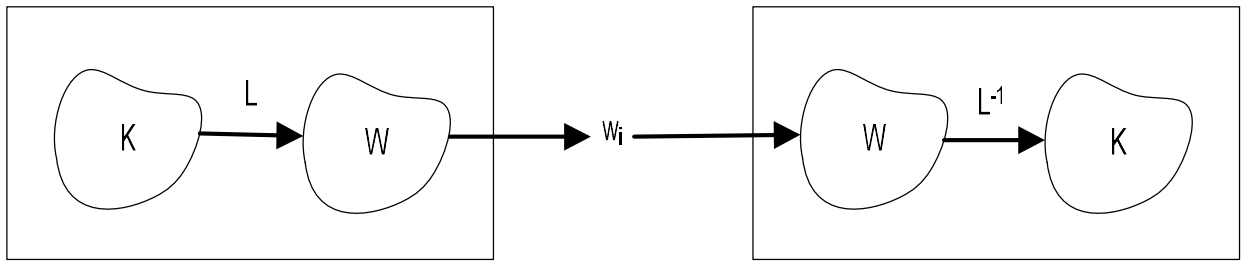

Figure 1. A transmitting agent (left) maps selected concepts abstract representations from set K into words of set W. These are received by a receiving agent (right) and mapped back into the concepts set K. Both sets K and W and the language functions have been learned and reside somewhere in the agents brains. This is an idealized but useful view (see text).

The inclusion (or exclusion) of a candidate concept within the set K depends upon the agent sensory capabilities, its instincts, interests and motivations, as well as its world or environment properties. We assume that agents who share the same nature, culture and environment, i.e. belong to the same community, will all learn an identical language function and its inverse, identical K and W sets and identical $P_{ij}$ conditionals matrix; this will rarely be strictly true since learning by individual agents will generally be imperfect and approximate but is a useful simplification.

It may be argued (e.g. Havlik, 2024) that the basic element of the set of abstract representations is a *thought* (rather than a concept as depicted above) and that the corresponding basic vocabulary element is a *sentence* (rather than a word). This may be (or may be not) true. However, adopting this approach will unnecessarily burden our notation, with no significant change in our arguments and conclusions.

3. THE LARGE LANGUAGE MODELS CASE

In contrast with the above described agents, LLMs do *not* learn the functions L and $L^{-1}$ *nor* learn the set K. Instead, the vocabulary set W is provided to the LLM a-priori (typically in the form of parts of words, so called tokens, for implementation efficiency reasons) and the grand matrix $P_{ij}$ is set as objective function and learned in a self-supervised fashion by presenting it with a large corpus (originally created by intelligent and communicative agents) of a selected language and repeatedly asking it to predict the next (or a masked) word.

The extraordinary fluency, human-like quality of LLMs generated dialogues, and their apparent reasoning capabilities have led many to believe or support the idea (as referred to in Section 1 above) that these models 'know what they talk about' and are not just 'stochastic parrots' (Bender et al, 2021). A leading argument in favor of this thesis is the "you shall know a word by the company it keeps" maxim (Firth, 1957), a motivating statement in distributional semantics.

We intend to provide a semi-formal proof that this thesis is false: LLMs, unlike humans and other hypothetical intelligent and communicative agents, do not, and cannot, have any understanding of the meaning of their dialogues.

We will make use of an ambiguity argument, repeatedly exploiting the trivial but profound truism: "*what exists, is possible*" (Boulding, 1957).

To prove our claim we turn to a simple thought experiment. Consider a pair of very different agents living in communities of very different worlds. These could be, for example, two humans (e.g. an Amazonian Waorani and an Alaskan Inuit); or a human and a hypothetical intelligent and communicative ant; or a human and an intelligent creature living in some other planet. The first agent (in any of these pairs) will have learned its own K set of concepts, its own vocabulary W, its own language map L (and its inverse), and its conditionals matrix $P_{ij}$. The second, will have learned its own K', W', L' (and its inverse) and $P_{ij}$'. In particular we may conceive a situation where, because they are living in extremely disparate worlds, their concepts sets K and K' are disjoint: no concept is shared amongst them. Their world views are totally different.

We now apply for a first time our "*what exists, is possible*" truism: since a concept set K of size N *exists* we must accept the *possibility* that our second (disjoint) set, K', may also be of same size. We assume then |K| = |K'| = N. Recall, we are dealing with possibility, no matter how small the probability might be; and just because some community will have developed a concept set of size N should not prevent any other from doing it as well. Now, since L is a bijective map (Eq. 1 above) we also immediately have |W| = |W'| = N. We proceed by applying this same exists/possible reasoning a second time and similarly conclude that the possibility exists for conditional probability matrices to be (approximately) equal. We cannot expect more than approximation to some arbitrary error here since these matrices values are real numbers, but for our purpose we may consider them equal. We will therefore assume hereon that $P_{ij} \cong P_{ij}'$.

One step further, recalling as mentioned above that the realization of the vocabulary W into utterances or textual symbols $w_i$ is arbitrary and jointly agreed by convention by community members, we may update all the elements of W' s.t. $w_i' = w_i \; \forall \; i = 1, 2, …N$ with no material impact.

To summarize, we now have two communities in different worlds with two disjoint concepts sets which are of equal size, equal vocabularies and equal conditional probability matrices.

Next, returning back to our LLM, we let it learn the conditionals matrix $P_{ij}$ (under the preset vocabulary W) of, say, our first agent. As per our LLM training protocol, briefly described above, neither the concepts set K nor the language map L are learned during this process.

After completion of $P_{ij}$ learning we may ask our LLM what the meaning of any word $w_m \in W$ is. But this question meets an inherently ambiguous situation: having $P_{ij} \cong P_{ij}'$ and W = W', *there is no way* for our LLM to resolve whether $w_m$ refers to a concept belonging to the first world set K or to a totally different concept belonging to the second world concepts set K'. Even assuming our LLM can assign meaning to any word (and not only to e.g. homonyms which we ignore herein as mentioned above) based on "the company it keeps" it would have no clue which one to select, whether meanings from K or meanings from K'. Note that an intelligent and communicative agent, having concurrently learned not only $P_{ij}$ and W but also K and L (both of its own world) has no such ambiguity problem.

Hence, there seems to be no alternative except concluding that the words learned by an LLM are totally detached from their concepts and, as a corollary, from their meanings.

4. DISCUSSION

We believe to have provided in this brief (and modest) paper a semi-formal proof that LLMs cannot really understand the meaning of their dialogues, no matter how intelligent, fluent and sophisticated these might seem to us, humans.

It has been said (e.g. Elman, 1990, amongst others) that humans cannot learn the meanings of a language just by listening to the radio (or, equivalently, just by reading texts). Since 'reading texts' is the only activity during LLMs training process, our conclusion that LLMs cannot achieve language understanding should not be a surprise to us.

Understanding the world is a prerequisite for (and not a result of) understanding language. As it may be evident by now world understanding apparently requires the existence, within organisms' brains or LLMs architecture, of a set of concepts abstract representations, including their mutual relations. Such a set is evidently missing in any current LLM architecture. In contrast such a set exists in human brains as proved by (Fried et al., 1997), and should most probably be found (should appropriate experimental techniques be developed) also within the brain of any intelligent, not necessarily communicative, organism.

Those supporting the thesis that LLMs do understand language may argue that languages are not such simplistic entities as we have been assuming herein, and that their fine-grain structure which includes syntax rules (along with long lists of exceptions), synonyms, homonyms, etc. really matter with regards to meaning acquisition. For example, they may claim, decision on the selected meaning of a homonym exclusively depends on "its company" (and we would agree on this). However, we would be very surprised at the same time, if these language idiosyncrasies, which we ignored in this paper, should be accredited to be the pillars of LLMs comprehension.

The false belief in the ability of LLMs to understand language may be after all, as has been already suggested by (Bender and Koller, 2020; Sejnowski, 2022; and others) a 'theory of mind' reflection of our own understanding of what an LLM says – that is a case of the illusionary so called Eliza (Weizenbaum, 1966) effect. In our thought experiment above, our intelligent and communicative ant and a human, when both listening to an identical LLM discourse, will experience each a distinct Eliza effect, each interpreting the message in accordance with his own set of world concepts, completely different from each other.

Finally, if progress toward human-like world and language understanding is desired we should continue our search, which first step should include a major, yet undiscovered, paradigm shift.